%% file: tmi.tex
\documentclass[journal,twoside,web]{ieeecolor}
\usepackage{tmi}
\usepackage{cite}
\usepackage{amsmath,amssymb,amsfonts}
\usepackage{algorithmic}
\usepackage{graphicx}
\usepackage{textcomp}
\usepackage{algorithm}

\usepackage{adjustbox}
\usepackage{threeparttable}
\usepackage{booktabs}
\usepackage{multirow}
\usepackage{tabularx}
\usepackage{subfigure}

\usepackage{hyperref}
\hypersetup{hidelinks=true}

\def\BibTeX{{\rm B\kern-.05em{\sc i\kern-.025em b}\kern-.08em
    T\kern-.1667em\lower.7ex\hbox{E}\kern-.125emX}}
\begin{document}
\title{Interpretable Multimodal Cancer Prototyping with Whole Slide Images and Incompletely Paired Genomics}
\author{Yupei Zhang, Yating Huang, \IEEEmembership{Student Member, IEEE}, Wanming Hu, Lequan Yu, \IEEEmembership{Member, IEEE}, Hujun Yin, \IEEEmembership{Senior Member, IEEE}, Chao Li
\thanks{This work was supported by the Guarantors of Brain. (Yupei Zhang and Yating Huang contributed equally to this work.)}
\thanks{Y. Zhang is with Department of Clinical Neurosciences, University of Cambridge, UK (e-mail: yz931@cam.ac.uk).}
\thanks{Y. Huang is with Department of Electrical \& Electronic Engineering, The University of Manchester, UK (e-mail: yating.huang@manchester.ac.uk).} 
\thanks{W. Hu is with Department of Pathology, State Key Laboratory of Oncology in South China, Guangdong Provincial Clinical Research Center for Cancer, Sun Yat-sen University Cancer Center, China (e-mail: huwm@sysucc.org.cn).} 
\thanks{L. Yu is with Department of Statistics and Actuarial Science, The University of Hong Kong, Hong Kong SAR, China (e-mail: lqyu@hku.hk).}
\thanks{H. Yin is with Department of Electrical \& Electronic Engineering, The University of Manchester, UK (e-mail: hujun.yin@manchester.ac.uk).}
\thanks{C. Li is with Department of Clinical Neurosciences and Department of Applied Mathematics and Theoretical Physics, University of Cambridge; School of Science and Engineering and School of Medicine, University of Dundee, UK (Corresponding author, e-mail: cl647@cam.ac.uk.)}
}

\maketitle

\begin{abstract}

Multimodal approaches that integrate histology and genomics hold strong potential for precision oncology. However, phenotypic and genotypic heterogeneity limits the quality of intra-modal representations and hinders effective inter-modal integration. Furthermore, most existing methods overlook real-world clinical scenarios where genomics may be partially missing or entirely unavailable. We propose a flexible multimodal prototyping framework to integrate whole slide images and incomplete genomics for precision oncology. Our approach has four key components: 1) Biological Prototyping using text prompting and prototype-wise weighting; 2) Multiview Alignment through sample- and distribution-wise alignments; 3) Bipartite Fusion to capture both shared and modality-specific information for multimodal fusion; and 4) Semantic Genomics Imputation to handle missing data.
Extensive experiments demonstrate the consistent superiority of the proposed method compared to other state-of-the-art approaches on multiple downstream tasks. The code is available at \href{https://github.com/helenypzhang/Interpretable-Multimodal-Prototyping}{GitHub}.

\end{abstract}

\begin{IEEEkeywords}
Multimodal Learning, Prototype Learning, Computational Pathology, Missing Modality
\end{IEEEkeywords}

\section{Introduction}
\label{sec:introduction}

\IEEEPARstart{P}{recision} oncology increasingly relies on multimodal approaches that provide complementary information. Whole slide images (WSIs) capture tissue morphology and cellular phenotypes, whereas genomics probes underlying molecular mechanisms. Integrating both offers improvement in cancer diagnosis, prognosis, and treatment stratification with phenotypic and genotypic characteristics \cite{chen2020pathomic, song2024multimodal, zhou2023cross}. 

Due to the high dimensionality and inherent disparity of WSIs and genomics, challenges remain for effective multimodal learning:
\textbf{1) Suboptimal WSI representation:} 
WSIs are gigapixel-scale images that typically lack detailed patch-level annotations \cite{xiang2022dsnet}, resulting in limited label or semantic information for representation learning. To address this, most existing methods employ multiple instance learning (MIL), where patch-level embeddings are aggregated into slide-level. However, these methods treat patches as independent instances \cite{bilal2023aggregation}, thereby ignoring key cancer phenotypes, such as spatial tissue organization that reflects tumor microarchitecture and cellular interactions. 
Although recent advances introduce attention-based MIL \cite{shao2021transmil} that models patch dependencies, aiming to focus on the most relevant regions. Yet, attention mechanisms may inadvertently highlight non-diagnostic or texture-based regions rather than pathology-relevant areas \cite{liu2024attention, albuquerque2024characterizing}. This results in suboptimal WSI representations failing to capture clinically meaningful patterns, hence limiting downstream multimodal integration with genomics.


\textbf{2) Ineffective cross-modal alignment due to semantic gaps:} 
Histology and genomics describe biological processes at distinct levels, resulting in substantial semantic discrepancies between feature spaces. 
Aligning distinct modalities before integration is crucial for multimodal learning \cite{li2021align}. 
Contrastive learning shows promise for cross-modal alignment \cite{radford2021learning}, encouraging paired WSI-genomic representations to converge while separating unpaired ones \cite{vaidya2025molecular}. However, in most clinical datasets, patients have multiple WSIs but only a single genomic profile \cite{cooper2018pancancer}. This imperfect pairing introduces ambiguity and noise in alignment, limiting the discriminative power of learned representations. Constructing robust and coherent cross-modal embeddings remains a challenge.

\textbf{3) Coarse integration strategies that overlook modality-specific features:} 
Effectively integrating WSIs and genomics requires not only capturing their shared biological signals (e.g., proliferation or metabolics),  but also preserving modality-specific signals (e.g., spatial context in histology or mutations in genomics) \cite{zhang2024prototypical}. 
However, existing approaches either embed both modalities into a unified latent space or focus solely on modeling global correlations \cite{schouten2025navigating}. While this facilitates fusion, it often suppresses modality-unique information and limits interpretability and downstream performance. 
To harness the full potential of cross-modal complementarity, a fine-grained integration is required. 
However, achieving this balance is non-trivial as it requires disentangling modality-specific signals from shared representations while minimizing redundancy.

\textbf{4) Limited robustness under incomplete genomics:} 
In real-world clinical scenarios, genomic data are often incomplete, inconsistently acquired, or entirely absent due to time constraints, tissue availability, or sequencing costs \cite{xing2022discrepancy}. 
Most existing methods \cite{chen2021multimodal, zhou2023cross} typically assume complete data availability across modalities or distill histology-only models to address incompleteness \cite{xing2022discrepancy}. 
However, they lack flexibility in handling heterogeneously available genomics, partially or entirely missing genomics, in a unified framework, limiting their generalizability and clinical deployment.

To address these challenges, we propose a multimodal learning framework integrating WSIs and incompletely paired genomics. \textbf{Firstly}, a \textbf{Biological Prototyping} scheme is proposed to model interpretable intra-modal representations. 
For WSIs, we leverage pathology domain-informed text prompts to derive meaningful visual histology prototypes from unannotated WSI patches. For genomics, we introduce functional grouping based on known biological functions to construct genomic prototypes. A prototype-wise weighting mechanism is further proposed to assign importance scores to prototypes, enhancing explainability and allowing models to focus on key modality-specific features. 
\textbf{Secondly}, a \textbf{Multiview Alignment} strategy is proposed to bridge semantic gaps across modalities.  
We first introduce a mutual information maximization loss to ensure distribution-level alignment and global semantic consistency. To alleviate imperfect pairing in real-world multimodal datasets, we further propose a sample-wise alignment that captures inter-sample relational consistency across mini-batches. Together, these strategies facilitate robust and complementary geno-phenotypic representations. 
\textbf{Thirdly}, we propose a \textbf{Bipartite Fusion} that captures both shared and unique features across modalities by leveraging cross-modal affinity, enabling fine-grained semantic fusion while maintaining modality-specific information, thereby enhancing both performance and interpretability.
\textbf{Finally}, to improve robustness in the presence of missing genomics, we design a \textbf{Semantic Genomics Imputation} module. This module flexibly handles various missingness scenarios, including fully missing, partially missing, or absent during inference, thereby enhancing clinical applicability.
Extensive experiments on multiple cancer downstream tasks in three datasets demonstrate that our method outperforms other state-of-the-art (SOTA) methods.

\section{Related Work}
\subsection{Multimodal Learning for Precision Oncology}
Recent multimodal approaches with WSIs and genomics primarily focus on two aspects: intra-modal representations and inter-modal integrations.
For WSIs, traditional approaches employed MIL to aggregate patch-level features into slide-level representations. However, these methods typically neglect inter-patch correlations and lack biological interpretability, limiting their clinical utility.
For genomics, recent studies \cite{liberzon2015molecular, chen2021multimodal, zhou2023cross} employ a coarse categorization into six genomics families; others \cite{song2024multimodal, jaume2024modeling} utilize biological pathways for transcriptomic representation. 
Different from these studies, we propose a biological prototyping scheme that simultaneously achieves two critical objectives: (1) capturing biologically relevant patterns in WSI patches and genomic tokens to form modality-specific prototypes, and (2) dynamically quantifying prototype importance through predicted weights to enable model interpretation at both phenotype and genotype levels.

To integrate histology and genomics, earlier studies integrated WSI Region Of Interests (ROI) and genomics profiles \cite{chen2020pathomic, xing2022discrepancy, pan2024focus}.  
Although straightforward, ROIs are not readily available in real-world clinical scenarios. 
Recent studies have focused on WSIs and transcriptomes \cite{song2024multimodal, jaume2024modeling}. For instance, SurvPath \cite{jaume2024modeling} introduced biological pathways for transcriptomic modelling and integrated WSIs to predict patient survival. 
However, the inherent semantic gap between genotype and phenotype hinders multimodal fusion, ultimately degrading model performance. Unlike conventional approaches, our method employs: \textbf{(i)} Multiview Alignment to reduce semantic discrepancies through cross-modal correlation learning, and \textbf{(ii)} a Bipartite Fusion module that dynamically enhances discriminative features while suppressing noise, enabling fine-grained prototypes with enriched semantic representations.

\begin{figure*}[!t]
    \centering
    \includegraphics[width=\textwidth]{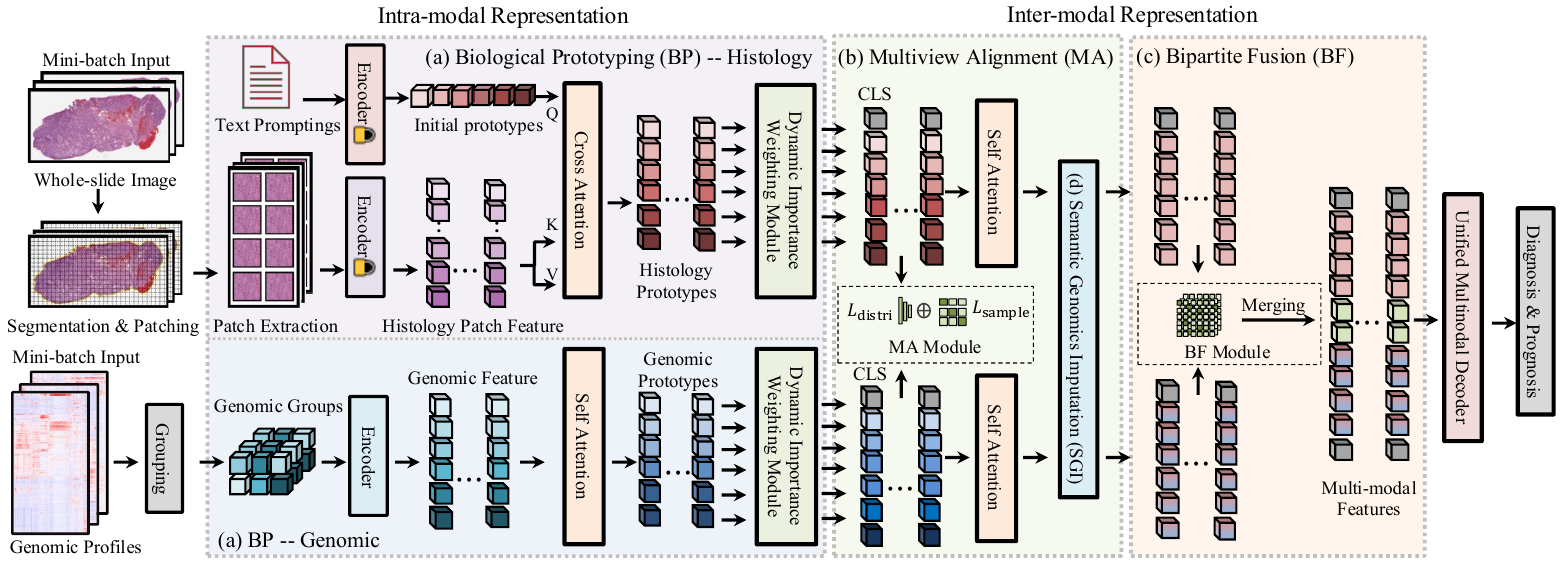}
    \caption{An overview of the proposed framework. (a) Biological Prototyping (BP): Extracts biologically meaningful prototypes from WSI and genomic data through class-specific promptings and a dynamic importance weighting module, enhancing feature interpretability. (b) Multiview Alignment (MA): Bridges the semantic gap between WSI and genomic features using sample-wise alignment and mutual information maximization. (c) Bipartite fusion (BF): Integrates shared and modality-specific features to optimize multimodal fusion. (d) Semantic Genomics Imputation (SGI): Handles missing genomic data scenarios by imputing features directly in the feature space. 
    }\label{fig:framework}
\end{figure*}

\subsection{Foundational Multimodal Pretraining}
\label{sec:foundation}

Recently, foundational multimodal pretraining has emerged in computational pathology, enabling the integration of large-scale, unlabeled, and unpaired multimodal data for task-agnostic model learning \cite{lu2023visual}. These models leverage cross-modal information, typically image, text, and omics, to learn transferable representations applicable to various downstream tasks. In vision-language pretraining, PLIP \cite{huang2023visual} utilized 208,414 paired pathology images and natural language descriptions from medical social media to perform contrastive language-image pretraining. CONCH \cite{lu2024visual} extended this by training on 1.17 million image-caption pairs of WSI and biomedical text using contrastive learning. PathChat \cite{lu2024multimodal} integrated a foundational pathology vision encoder with a large language model and fine-tuned on over 456,000 visual-language instruction pairs involving 999,202 question-answer interactions, to function as a generalist AI assistant in human pathology. MUSK \cite{xiang2025vision} pretrained a unified masked modeling architecture on one billion pathology-related text tokens and 50 million pathology images of 11,577 patients, followed by alignment with one million paired image-text examples. 

\subsection{Multimodal Learning with Incomplete Modality}
The absence of genomic data poses a critical challenge to multimodal learning, as acquiring comprehensive genomics profiles remains expensive and relies on tissues \cite{pan2024focus, wang2025histo}.
To tackle this, Xing et al. \cite{xing2022discrepancy} transferred multimodal teacher knowledge to a uni-modal student model by a distillation framework for glioma grading. Wang et al. \cite{wang2023multi} designed a multi-task framework that jointly predicted molecular status and glioma classification from WSIs, requiring genomics only during training.
Despite promising, these methods still rely on paired multimodal data during training. Recent methods \cite{qiu2024dual} proposed imputation-based methods to leverage incomplete histology and genomics during training.
However, the substantial domain gap between histology and genomics limits the effectiveness of direct imputation. To address these limitations, we proposed a semantic genomics imputation module, enabling our model to adapt to various real-world scenarios.

\section{Methodology}
Fig.~\ref{fig:framework} provides an overview of the proposed multi-modal prototype learning approach designed to model intra- and inter-modal representations under WSIs and incompletely paired genomics. The proposed framework includes: 1) a prompting-based Biological Prototyping module for interpretable intra-modal representation (Section \ref{sec:A}); 2) a Multi-View Alignment module (Section \ref{sec:B}) and Bipartite Fusion (Section \ref{sec:D}) for inter-modal representation; 3) a Semantic Imputation module to handling incomplete genomics (Section \ref{sec:C}). Section \ref{sec:E} shows the algorithm workflow.

\subsection{Biological Prototyping}
\label{sec:A}

To construct biologically meaningful and interpretable intra-modal representations, we propose the Biological Prototyping (BP) module (Fig.~\ref{fig:framework}(a)). BP initializes histology prototypes using pathology-specific text prompts and genomic prototypes using established gene functional groups, grounding both modalities in meaningful biological concepts. These prototypes serve as semantic anchors that guide feature learning toward recognizable histological and molecular patterns. A dynamic importance weighting mechanism further quantifies each prototype’s contribution to downstream tasks, enhancing transparency and interpretability.


\subsubsection{Histology prototypes}
We define six coarse-grained histological categories: \textit{Neoplastic, Necrotic, Inflammatory, Stromal, Infiltrative}, and \textit{Other Cell Types}, following prior work~\cite{gamper2020pannuke}. Initial prototype embeddings for each category are obtained using a pre-trained pathology-specific text encoder. To adapt prototypes to tumor heterogeneity and contextual variations within WSIs, we refine them using a cross-attention mechanism where patch embeddings act as keys/values and initial prototypes as semantic queries:
\begin{equation}
P^{t} ={\rm softmax}\left ( \frac{P^{t-1}W_{q}\left ( P_{\rm patch}W_{k} \right )^{T}  }{\sqrt{D} }  \right ) \left ( P_{\rm patch}W_{v} \right )
\label{eq1}
\end{equation}
where $P^{t}$ denotes updated prototypes at iteration $t$-th and $P^{t-1}$ represents histology prototypes at the $\left ( t-1 \right ) $-th interation. This mechanism encourages prototypes to evolve into semantically meaningful cluster centers. 
\subsubsection{Genomic prototypes}
Following established molecular taxonomies \cite{chen2021multimodal}, genes are grouped by their functional roles: \textit{Tumor Suppressor Genes, Oncogenes, Protein Kinases, Cell Differentiation Markers, Transcription Factors}, and \textit{Cytokines and Growth Factors}. Each group forms a genomic prototype encoding functionally coherent information. These prototypes are refined via a self-attention mechanism, modeling genotype-genotype dependencies. This allows genomic prototypes to adaptively represent salient molecular patterns.

\subsubsection{Dynamic importance weighting}
To explicitly compute the contribution of individual prototypes toward downstream tasks, we introduce task-specific importance scores $w$ for each prototype:
\begin{equation}
w_{i}=\sigma \left ( f_{\theta } \left ( z_{i} \right ) \right ) 
\label{eq2}
\end{equation}
where $\sigma$, $f_{\theta }$, and $z_{i}$ denote the sigmoid function, linear layers, and $i$-th prototype token, respectively. This mechanism allows the model to highlight informative features while suppressing irrelevant ones, improving interpretability and stability.



\begin{figure}[!t]
    \centering
    \includegraphics[scale=.55]{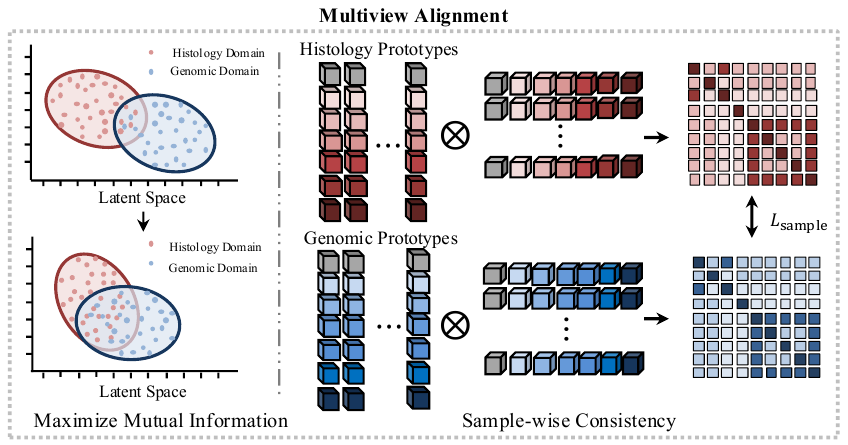}
    \caption{Multiview Alignment (MA): Bridges the semantic gap between WSI and genomic features using mutual information maximization and sample-wise alignment. Left: Maximize mutual information in MA. Right: Sample-wise consistency.}\label{fig:MA}
\end{figure}

\subsection{Multiview Alignment}
\label{sec:B}
WSIs and genomics inherently exhibit semantic gaps between modalities. Multimodal cancer datasets often contain imperfectly paired samples, which exacerbates the modality discrepancy.
To mitigate these challenges, we propose the Multiview Alignment (MA) module as shown in Fig.~\ref{fig:framework}(b) and Fig.~\ref{fig:MA}, which aligns the two modalities at both sample-wise and distribution-wise levels. 
Specifically, distribution-wise alignment enhances global cross-modal coherence by maximizing mutual information (MI), while the sample-wise strategy calibrates alignment by cross-sample batch consistency. Jointly optimizing these two alignment strategies effectively bridges modality gaps before the final fusion.

\subsubsection{Distribution-wise Alignment via Mutual Information Maximization}
MI quantifies the shared information between two modalities, serving as a robust measure of cross-modal dependency. 
As illustrated in Fig.~\ref{fig:MA}, we maximize MI to enforce distributional alignment between histology and genomic prototypes. 
For a mini-batch paired histology $P\in\mathbb{R}^{B\times N_P\times D_P}$ and genomic embeddings $G\in\mathbb{R}^{B\times N_G\times D_G}$ ($B$ as the batch-size, $N$ as the number of prototypes, D as the feature dimension), MI between modalities is defined as $I(P, G)=H(P)-H(P|G)$, $H(P)$ and $H(P|G)$ denote the entropy and conditional entropy).
Directly computing MI is challenging due to the intractability of high-dimensional probability distributions. To address this, we leverage a contrastive-based estimator:
\begin{equation}
L_{\rm MIE} =-\sum_{i=1}^{B} {\rm log}\frac{{\rm exp}\left ( f_{\rm MI}\left ( p_{i} , g_{i} \right )   \right ) }{ {\textstyle \sum_{i=1}^{B}{\rm exp}\left ( f_{\rm MI} \left ( p_{i} , g_{i} \right )  \right ) } }
\label{eq5}
\end{equation}
where $f_{\rm MI}\left ( p_{i}, g_{i} \right )$ is a learnable function estimating the mutual information score between $i$-th paired histology-genomic embedding $p_i\in\mathbb{R}^{N_P\times D_P}$ and $g_i\in\mathbb{R}^{N_G\times D_G}$. 

However, maximizing MI alone may lead to representation collapse, where embeddings become overly similar and lose representational diversity. To alleviate this issue, we introduce an entropy-based regularization term:
\begin{equation}
L_{\rm{reg}} = \frac{1}{B^{2}} \sum_{i=1}^{B} \sum_{j=1}^{B} \exp\left(-\|g_i - g_j\|\right).
\label{eq6}
\end{equation}
This regularization prevents degenerate solutions and ensures that the aligned genomic representations preserve their intrinsic variability while maintaining cross-modal coherence. The overall distribution-wise alignment loss thus combines contrastive MI maximization with entropy-based regularization:
\begin{equation}
L_{\rm distri} = L_{\rm MIE} + \lambda_{\rm reg} L_{\rm reg},
\label{eq7}
\end{equation}
where $\lambda_{reg}$ controls the strength of the regularization.

\begin{figure}[!t]
    \centering
    \includegraphics[scale=.55]{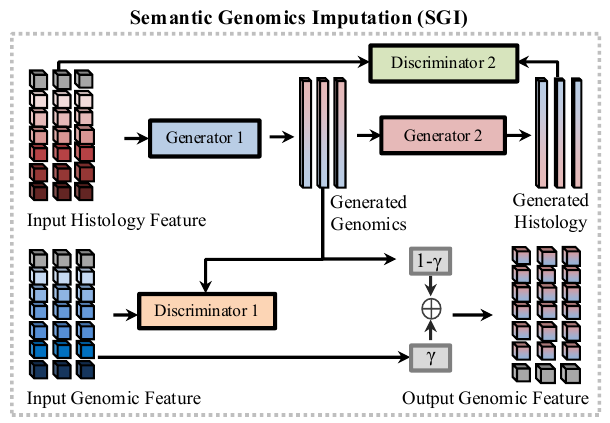}
    \caption{Semantic Genomics Imputation (SGI): A CycleGAN-based network to generate genomic features from histology features, with a progressive interpolation to ensure robustness in various settings.}\label{fig:SGI}
\end{figure}

\subsubsection{Sample-wise Alignment via Correlation Consistency}
Conventional contrastive-based alignment approaches treat each WSI-genomic pair as an independent positive sample, inadvertently pushing WSIs from the same patient apart as negatives, leading to suboptimal representations.
To mitigate this, we align relational structures across modalities rather than individual embeddings.
Given $P$ and $G$ as histology and genomic feature matrices within a batch of $B$ samples, we first apply L2-normalization to obtain scale-invariant representations: $\tilde{P}$ and $\tilde{G}$.
We then compute their Gram matrices to capture intra-batch correlations:
\begin{equation}
M_P = \tilde{P}\tilde{P}^{\top}, \quad M_G = \tilde{G}\tilde{G}^{\top},
\end{equation}
where $M_P(i,j)$ and $M_G(i,j)$ represent the similarity between samples $i$ and $j$ within the histology and genomic modalities, respectively.
If two samples exhibit strong similarity in the genomic space, their corresponding WSIs should display comparable similarity patterns in the histology space. We therefore enforce relational consistency by minimizing the Frobenius norm between $M_P$ and $M_G$:
\begin{equation}
L_{\rm sample} = \frac{1}{B^{2}} \| M_P - M_G \|_{F}^{2},
\end{equation}
where $\|\cdot\|_{F}$ denotes the Frobenius norm. 
By aligning cross-modal representations at a structural level rather than individual embeddings, this design preserves intra-patient and inter-sample relationships across modalities, preserving biological coherence under imperfect WSI-genomic pairing.

Finally, the MA module combines both sample-wise and distribution-wise alignment losses:
\begin{equation}
L_{\rm MA}=L_{\rm sample}+L_{\rm distri}
\label{eq8}
\end{equation}
To capture inter-prototype interactions after alignment, we introduce a learnable $\left [ CLS \right ]$ token for each modality and apply a self-attention layer over all prototypes. This allows the model to encode their dependencies and produce a global contextual representation.

\subsection{Semantic Genomics Imputation}
\label{sec:C}
To handle incomplete genomic data in clinical practice, we propose a Semantic Genomics Imputation (SGI) module (Fig.~\ref{fig:SGI}) that infers missing genomic embeddings directly from histology features.
Benefiting from the pre-aligned feature space provided by MA, the imputed genomic features preserve biological fidelity, ensuring robust predictions even when genomics are partially or entirely absent.

Inspired by the cycle-consistency in CycleGAN~\cite{zhu2017unpaired}, SGI performs bidirectional feature-level translation between histology and genomic representations:
$F_{P\to G}$ translates histology embeddings into genomic embeddings, and
$F_{G\to P}$ reconstructs histology embeddings from genomic features.
The forward cycle consistency loss is defined as:
\begin{equation} L_{\rm cycle}^{P} = E_{p_{i}\sim P } \left \| F_{G\to P}\left ( F_{P\to G} \left ( p_{i}\right ) \right ) -p_{i} \right \|_{1} 
\label{eq9} 
\end{equation}
enforcing reconstruction of the original histology embedding after bidirectional translation.
A symmetric term $L{\rm cycle}^{G}$ is similarly applied for genomic embeddings under reverse transformation.

Beyond reconstructive consistency, it is essential that imputed genomic embeddings follow the true statistical distribution of real genomic features.
To this end, we incorporate an adversarial regularization strategy, where discriminators $D_{G}$ and $D_{P}$ are trained to distinguish between real and generated features in the genomic and histology embedding spaces, respectively. The adversarial loss for histology-to-genomics translation is defined as:
{\small \begin{equation} L_{\rm adv}^{G} =E_{g_{i}\sim G} \left [ logD_{G} \left ( g_{i} \right ) \right ] + E_{p_{i}\sim P}\left [ log\left ( 1-D_{G}\left ( F_{P\to G}\left ( p_{i} \right ) \right ) \right ) \right ] 
\label{eq11} 
\end{equation} 
}
A symmetric adversarial loss $L_{adv}^{P}$ is defined analogously for genomics-to-histology translation.
The final optimization objective integrates cycle consistency and adversarial learning:
\begin{equation}
L_{\rm FGI}= L_{\rm adv}^{G} +  L_{\rm adv}^{P} + \lambda \left (  L_{\rm cycle}^{G}+ L_{\rm cycle}^{P} \right )  
\label{eq13}
\end{equation}
where $\lambda$ controls the relative importance of cycle consistency loss and adversarial loss.

To stabilize training under varying levels of missingness, we adopt a progressive interpolation strategy that gradually shifts reliance from real to generated genomic embeddings:
\begin{equation}
\tilde{g_{i}} =m_{i}\cdot g_{i}+\left ( 1-m_{i} \right ) \cdot \hat{g_{i}} 
\label{eq14}
\end{equation}
where $g_{i}$ is a real genomic embedding, $\hat{g_i}$ is the generated one, and $m_{i}$ decreases over iterations. This smooth transition enhances robustness and allows the model to adapt seamlessly to different degrees of missing genomic data.
\begin{figure}[!t]
    \centering
    \includegraphics[scale=.55]{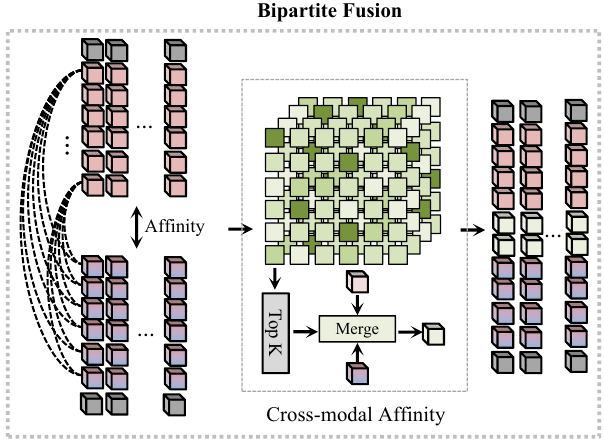}
    \caption{Bipartite Fusion: Utilizing prototype-wise affinity to integrate shared and modality-specific features for multimodal fusion.}\label{fig:BF}
\end{figure}

\subsection{Bipartite Fusion}
\label{sec:D}
To jointly preserve modality-specific semantics and capture cross-modal interactions, we propose a Bipartite Fusion (BF) module (Fig.~\ref{fig:BF}). 
In this design, histology and genomic prototypes form two disjoint node sets in a bipartite graph, where edges encode their pairwise affinities. 
Given the $i$-th paired histology and genomic prototype sets $p_i \in \mathbb{R}^{N_P \times D_P}$ and $g_i \in \mathbb{R}^{N_G \times D_G}$, the prototypes affinity is computed as:
\begin{equation}
A_i(n,m) = 
\frac{p_{i,n}^{\top} g_{i,m}}{\|p_{i,n}\| \, \|g_{i,m}\|},
\label{eq15}
\end{equation}
where $A_i \in \mathbb{R}^{N_P \times N_G}$ captures the affinity between the $n$-th histology prototype and the $m$-th genomic prototype.

We select the Top-$K$ most affine cross-modal prototype pairs $(p_{i,n}, g_{i,m})$ to perform cross-modal fusion, enabling localized and interpretable interactions between modalities. 
To retain modality-unique information, prototypes not participating in Top-$K$ correspondences are concatenated directly without fusion. 
The resulting representation for each sample $i$ thus integrates three complementary components: histology-specific, genomics-specific prototypes, and fused cross-modal prototypes. 
This design performs selective fusion of matched prototype pairs while retaining modality-specific features, supporting robust interpretability and fine-grained cross-modal representation learning.

\subsection{Algorithmic Pipeline}
\label{sec:E}
To ensure stable convergence, we adopt an alternating training pipeline that progressively introduces different objectives. 
The pipeline consists of two phases for each downstream task:
Step 1: Intra-modal initialization.
We first train the BP module independently with task objectives, enabling stable and discriminative intra-modal representation. 
Step 2: Inter-modal optimization.
Once intra-modal embeddings converge, adversarial learning is introduced to reconstruct missing genomics with a progressively increased missing rate during training, forcing the model to infer robust representations in various incomplete settings.
Meanwhile, the MA strategy is updated via batch accumulation.
Once MA achieves stable cross-modal coherence, the MA parameters are frozen to preserve modality balance achieved in earlier phases.
\input{tables/table1-2}

\section{Experiments \& Results} 

\subsection{Datasets and Implementations}
\subsubsection{Datasets}
We used two public datasets, TCGA-GBM and TCGA-LGG~\cite{tomczak2015review}, including both WSIs and genomic profiles for gliomas. Following studies \cite{lu2021data, wang2023multi}, we removed low-quality WSIs and samples without diagnostic annotations, retaining 939 patients and 1831 WSIs.
To evaluate cross-domain generalization, we included CPTAC-GBM \cite{li2023proteogenomic}, containing 177 glioblastoma patients with paired WSI and genomics.
\subsubsection{Implementations}
Our model was implemented in PyTorch and trained on NVIDIA Tesla A100 GPUs. We adopted the Adam optimizer with an initial learning rate of $1e^{-4}$ and with early stopping based on validation performance. We followed the standard MIL procedures, where each WSI was divided into non-overlapping patches, with patch-level features extracted using a pre-trained CLIP ViT-B/16 backbone. 

\input{tables/table3}
\subsection{Comparison with SOTA Methods}
We evaluated performance using AUC, accuracy, sensitivity, specificity, and F1-score for glioma diagnosis and grading, and concordance index (C-index) for survival prediction. All results were obtained using a five-fold cross-validation, and averaged over five random splits. We compared four uni-modal and seven multimodal SOTA methods. 
\input{tables/table4}

\subsubsection{Glioma Diagnosis and Grading}
As shown in Table~\ref{table1-2}, uni-modal histology models (e.g., AttMIL~\cite{ilse2018attention}, TransMIL~\cite{shao2021transmil}) perform suboptimally for both glioma diagnosis and grading, indicating that morphology alone is insufficient and that genomics provides essential complementary information.
In contrast, our method, even under the missing-modality setting (only WSI inputs), outperforms the WSI-only SOTA by $4.45\%$ and $3.79\%$ in AUC for diagnosis and grading, respectively. 
In the multimodal setting, naive fusion strategies such as addition or concatenation struggle to capture cross-modal dependencies, resulting in limited performance (e.g., Concat yields F1-score $61.26\%$ for grading). More advanced methods that explicitly model cross-modal interactions, such as HFBSurv~\cite{li2022hfbsurv}, MCAT~\cite{chen2021multimodal}, CMTA~\cite{zhou2023cross}, and MKD~\cite{zhang2025multi}, achieve stronger results, but still fall short of our approach, demonstrating the effectiveness of our inter-modal representation strategies.
\input{tables/table5}
\begin{figure*}[!t]
  \centering
  \includegraphics[scale=.40]{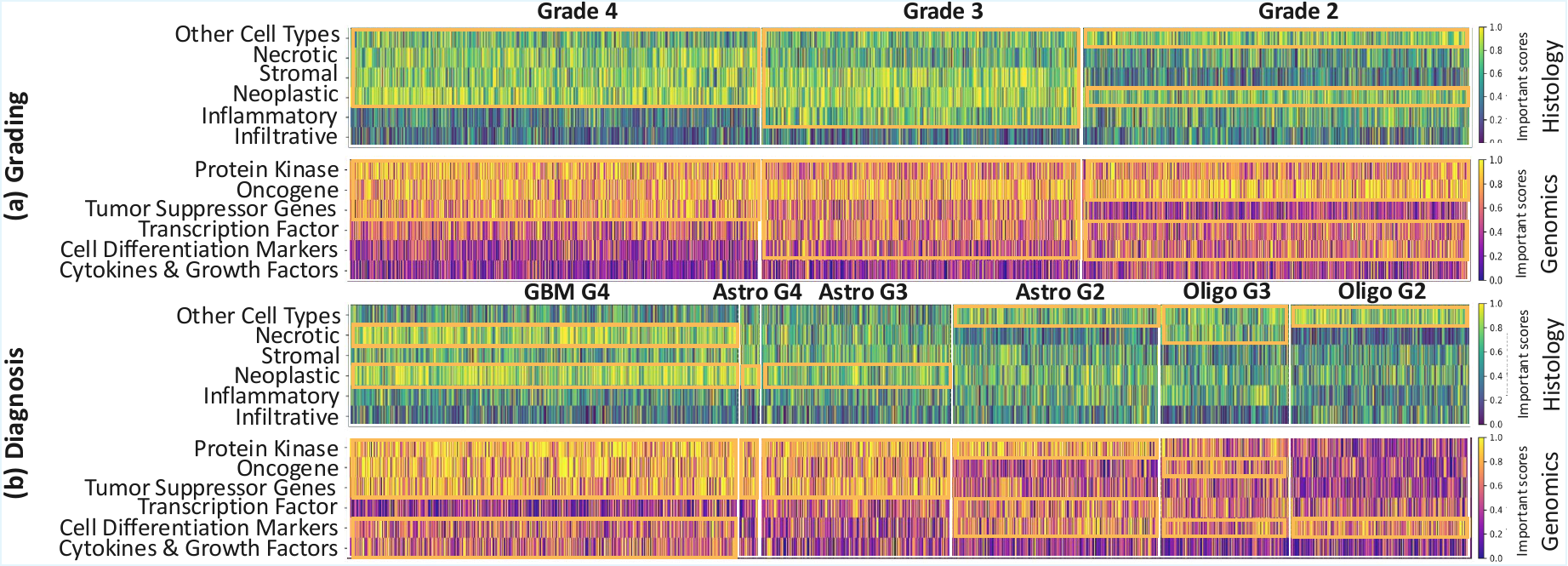}
  \caption{
    Interpretability heatmaps for glioma analysis across histology and genomics. 
    (a) Glioma Grading: patients are grouped in Grade 2-4.
    (b) Glioma Diagnosis (WHO 2021): GBM Grade 4, Astrocytoma Grade 4, Astrocytoma Grade 3, Astrocytoma Grade 2, Oligodendroglioma Grade 3, Oligodendroglioma Grade 2.
    In each task, upper heatmaps display histology prototypes (e.g., Neoplastic, Necrotic), while lower heatmaps illustrate the importance of genomic prototypes (e.g., Protein Kinase, Oncogenes). The importance score is normalized per patient. Our model demonstrates strong interpretability by identifying biologically meaningful markers for glioma grading and diagnosis.
  }
  \label{fig:vis_ih}
\end{figure*}

Overall, our model consistently performs the best across both diagnosis and grading. For diagnosis, our method attains the highest AUC (92.26\%), surpassing the second-best CMTA (90.47\%) by 1.79\%. For grading, our model achieves the best AUC (89.62\%), outperforming MCAT (88.26\%) by 1.36\%, while improving specificity by 2.40\% over CMTA. These consistent improvements demonstrate the effectiveness of our intra- and inter-modal representations, and the robustness under full and missing-modality, and cross-domain conditions.

\subsubsection{Survival Prediction}
As shown in Table~\ref{table3}, for survival prediction, our missing-modality model achieves a C-index of $75.95\%$, which is $2.23\%$ higher than the second-best uni-modal method (G-HANet~\cite{wang2025histo}), demonstrating its robustness for incomplete inputs. In multi-modal settings, existing methods such as MCAT and MKD obtain competitive in-domain results but exhibit limited cross-domain generalization. By contrast, our framework achieves the highest C-index both in-domain ($83.17\%$) and cross-domain ($64.01\%$) performance, highlighting improved generalization and distributional robustness of our model.
%

\subsection{Robustness to Genomic Modality Missingness}
To evaluate the robustness of our method in real-world scenarios with missing genomics, we tested two settings: 
\textbf{Patient-wise missingness} is defined as cases where genomics is entirely unavailable for certain patients, mirroring situations where sequencing cannot be performed due to constraints of costs, tissue or resources. 
\textbf{Feature-wise missingness} refers to scenarios where specific genes are absent (simulated in our experiments by randomly masking subsets of genes), simulating variability in the sequencing panel or experimental noise. Together, these settings enable a systematic assessment of the impact of complex missing scenarios on model performance.
We compared three representative strategies for addressing missingness using CMTA \cite{zhou2023cross} as the backbone.
The filling strategy replaces missing genomics with mean values. 
AE \cite{dumpala2019audio} trains an autoencoder to reconstruct genomic features from WSIs representations. 
The Ensemble approach trains separate models on each modality and aggregates their predictions.

As shown in Table~\ref{table4}, in the \textbf{patient-wise missingness} setting, baseline methods degrade in performance substantially as the proportion of missingness increases. 
The filling approach with mean values struggles to maintain predictive accuracy, as it fails to capture biological heterogeneity. The AE approach is limited in genomic representations due to substantial multi-modal semantic gaps. 
The ensemble approach is comparatively more stable by leveraging multiple modality-specific models; however, it fails to model cross-modal interactions, resulting in suboptimal performance. 
Our method consistently outperforms these baselines, as the proposed SGI module effectively reconstructs genomic embeddings from histology-derived features, ensuring biological coherence. 
These advantages are particularly evident at a higher missing ratio. 

A similar pattern is observed in feature-wise missingness, demonstrating the flexibility and robustness of our method. Compared to patient-wise missingness, the performance decline in \textbf{feature-wise} is generally more gradual, as models can still utilize the remaining genomics. 
Nonetheless, as the missing rate increases, the accumulation of noise adversely affects both filling and AE methods, resulting in performance degradation. Meanwhile, the ensemble approach, although relatively more stable, becomes sensitive to missingness. 
In contrast, our method performs well on all levels of feature-wise missingness, demonstrating consistent robustness.

Collectively, these findings highlight our advantages in maintaining biologically coherent representations, preserving cross-modal consistency, and robustly imputing incomplete genomics.  The strong performance under diverse missingness conditions underscores the translational reliability of our framework in real-world clinical scenarios where multimodal data is often incomplete.

\subsection{Ablation Study}
As shown in Table~\ref{table5}, we conducted comprehensive ablation studies of four proposed modules on all tasks under both missing and complete genomics. 
\subsubsection{Ablation of BP module}
We compared two initialization strategies: random initialization and CLIP text-based initialization using semantic histological and genomic labels. The text-based initialization substantially outperforms the random baseline, demonstrating that incorporating semantic priors not only improves discriminative capability but also enhances the interpretability through biologically coherent prototypes.

\subsubsection{Ablation of MA module}
We separately removed the sample-wise and distribution-wise alignment components. Ablating either degrades performance, while removing both leads to the most pronounced degradation. The sample-wise alignment improves robustness across both full and missing modality settings, while the distribution-wise alignment offers additional improvements, particularly when genomics are missing, as it regularizes latent feature dependencies and guides imputation. These findings confirm that the two alignment mechanisms are complementary and jointly crucial for robust cross-modal representation learning.
\input{tables/table6}
\begin{figure}[!t]
    \centering
    \includegraphics[scale=.35]{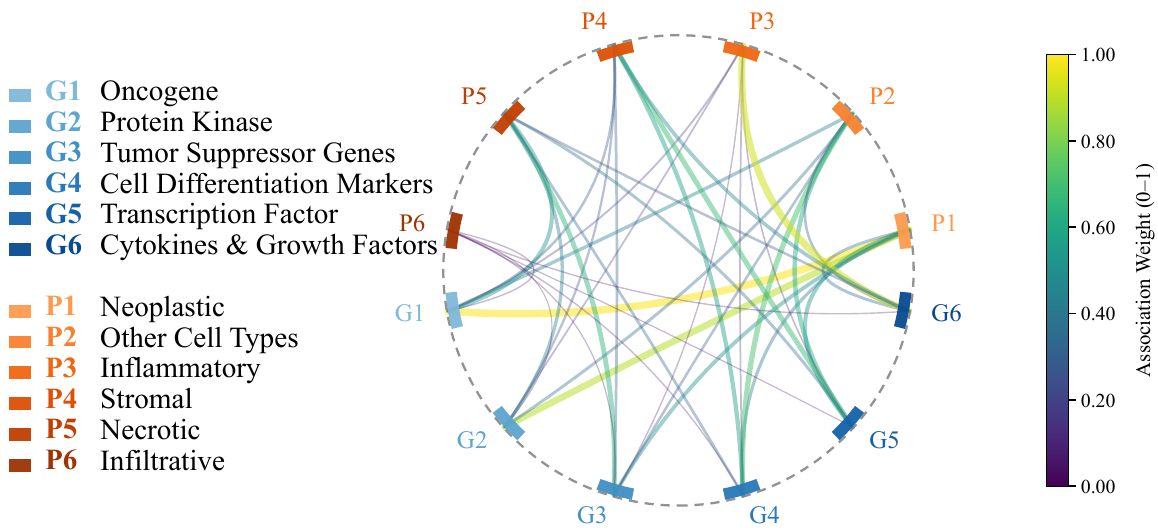}
    \caption{Visualization of cross-modal interactions between genomic prototypes (G1-G6) and histology prototypes (P1-P6) on survival prediction task. Edge color represents geno-phenotype association weights. }\label{fig:vis_cmi}
\end{figure}

\subsubsection{Ablation of SGI module}
Removing the SGI module forces the model to rely solely on paired WSI-genomic data, disabling its ability to handle missing modalities. A diffusion-based imputation variant underperforms, likely due to its limited capacity to model biologically meaningful cross-modal dependencies. Similarly, a GAN-based variant without the proposed progressive interpolation degrades sharply under partial missingness, underscoring the importance of gradually exposing incomplete genomics during training.
Moreover, even with all modalities available at inference, incorporating SGI during training consistently enhances performance. This can be attributed to the regularization effect of adversarial and reconstruction objectives, which strengthen cross-modal consistency and promote biologically coherent latent representations.

\subsubsection{Ablation of BF module} Replacing BF with a simple concatenation results in clear performance degradation, particularly under missing modality settings. This may be due to its ability to suppress noisy or redundant features arising from imputation, while retaining informative modality-specific signals. This selective fusion design proves more effective than naive aggregation, underscoring its importance in managing heterogeneous multimodal data.
\begin{figure}[!t]
    \centering
    \includegraphics[scale=.39]{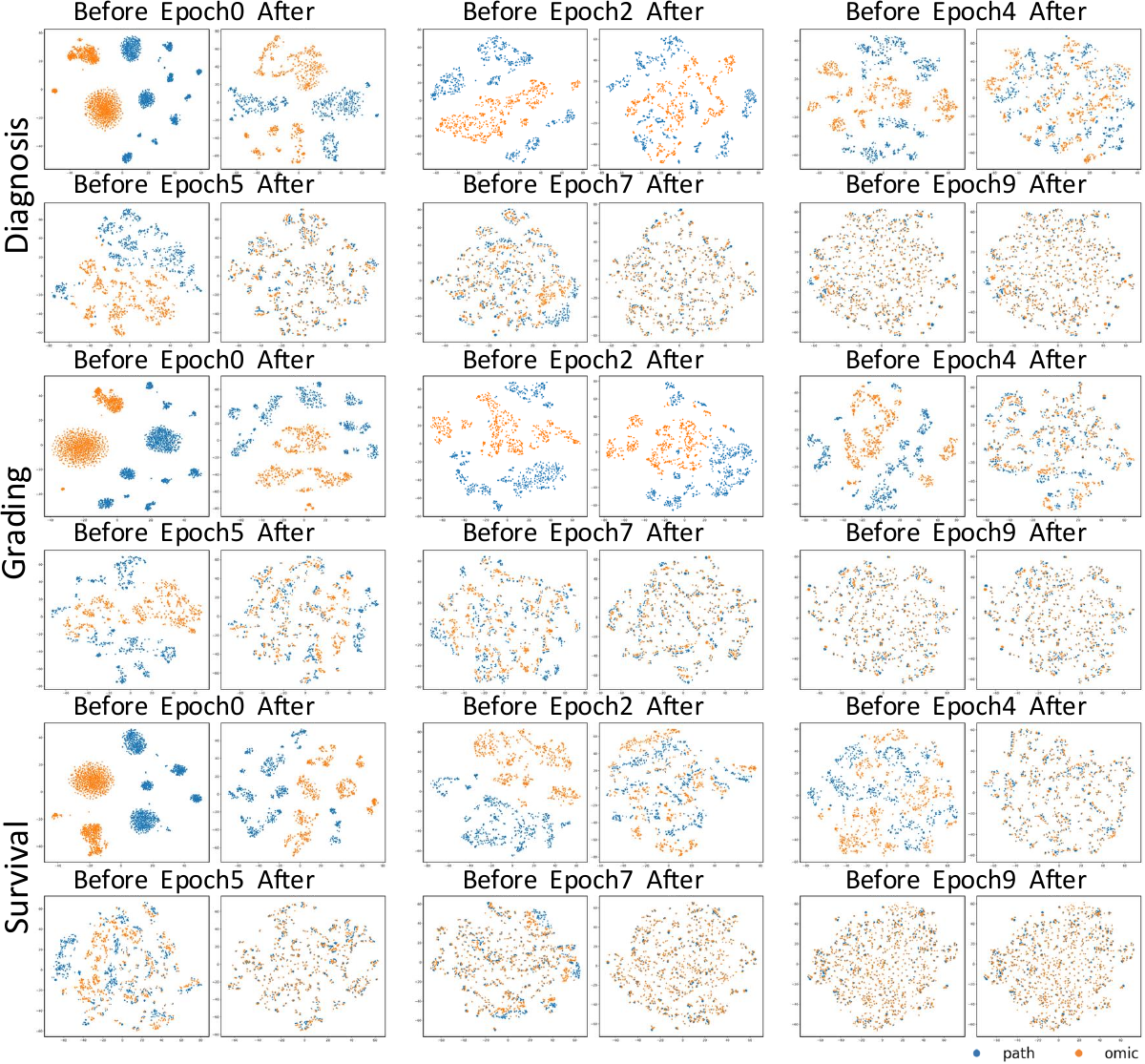}
    \caption{T-SNE visualization of histology and genomic features before and after multimodal alignment over 10 training epochs across three downstream tasks: diagnosis, grading, and survival prediction. Blue and orange points denote histology and genomics embeddings, respectively.}
    \label{fig:vis_tsne}
\end{figure}
\begin{figure*}[!t]
    \centering
    \includegraphics[scale=.50]{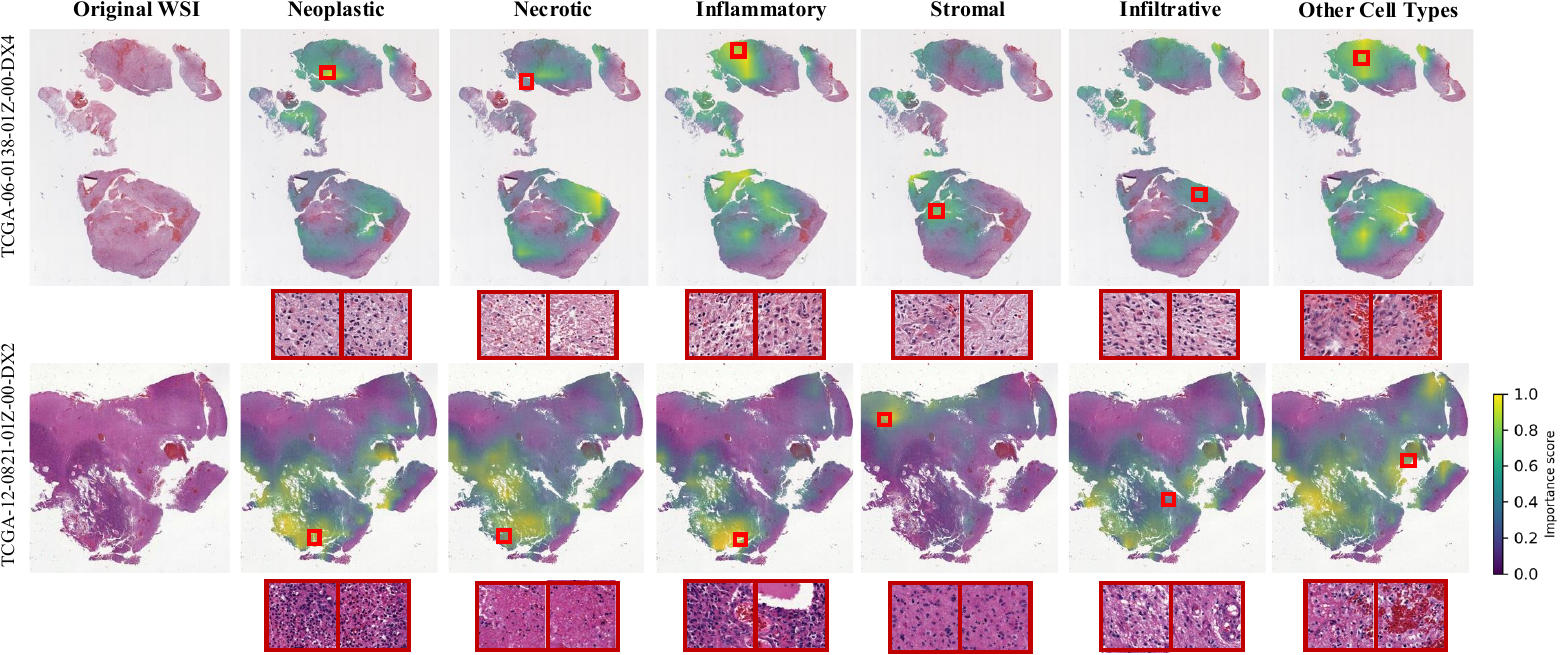}
    \caption{Spatial attention heatmaps of six histology prototypes on WSIs, showing phenotype-specific activation patterns for survival prediction.}
    \label{fig:vis_wsi}
\end{figure*}

\subsection{Analysis of Hyper-Parameters}
We further investigated the effect of the Top-$K$ parameter in the BF module on survival prediction. As described in Section~\ref{sec:D}, Top-$K$ controls how many cross-modal prototype pairs are fused, balancing integration and modality-specific preservation.  
Table~\ref{tab:ablate_k_survival} shows that performance remains relatively stable across different $K$. However, smaller $K$ values (e.g., $K=0,1$) underexploit cross-modal interactions, resulting in weaker performance in both in-domain and cross-domain settings. 
The setting $K=3$ achieves the highest in-domain accuracy (83.17\%), while also providing competitive cross-domain performance (64.01\%), only slightly below the best cross-domain result obtained at $K=2$ (64.88\%). 
In terms of computational efficiency, both $K=2$ and $K=3$ reduce training time compared to smaller values, likely because fusing similar pairs suppresses redundant information and accelerates convergence. 
Overall, $K=3$ provides the best balance between in-domain performance, generalization and training efficiency, and is thus adopted as the default configuration.

\subsection{Interpretability Analysis}
\subsubsection{Prototype-level Interpretability}
To investigate biological patterns learned by our model, we visualize multimodal importance contributions across patients in Fig.~\ref{fig:vis_ih}. Each column represents a patient, grouped by clinical labels, and rows refer to proposed prototypes derived from WSIs or genomics. Color intensity denotes the relative importance score of each prototype.

In Fig.~\ref{fig:vis_ih} (a), Grade 4 gliomas show high reliance on oncogenes and protein kinases in genomic prototypes, consistent with high-grade tumor hallmarks such as cell proliferation and immune escape \cite{tomuleasa2024therapeutic}. Also, their histological profiles emphasize neoplastic and necrotic components, characteristics of aggressive tumors. 
Grade 3 exhibits an intermediate dependence on transcription factors and tumor suppressor genes, accompanied by a greater emphasis on inflammatory histology. 
Grade 2 is characterized by cell differentiation marker genes and relatively normal-appearing tissue in histology, indicating a more indolent phenotype.

Fig.~\ref{fig:vis_ih} (b) further shows that the model differentiates glioma subtypes by capturing subtype-specific genomic and histology signatures. 
For example, GBM (Grade 4, IDH-wildtype) prominently emphasizes oncogenic signaling and necrotic morphology, whereas astrocytomas exhibit a grade-dependent transition from prominent oncogene signaling (in Grades 4 and 3) to increased involvement of transcriptional regulation and differentiation (in Grade 2). Oligodendrogliomas, particularly in lower grades, demonstrate consistent importance of cell differentiation markers and histology of other cell types.
Overall, the learned prototypes exhibit coherence with prior knowledge across both modalities and tasks, suggesting their relevance to underlying biology.

\subsubsection{Cross-modal Prototypes Interaction}
Fig.~\ref{fig:vis_cmi} visualizes the learned cross-modal interactions for the survival prediction task using a chord diagram. Genomic prototypes (G1-G6) and histology prototypes (P1-P6) are represented as nodes, with edges reflecting the magnitude of their associations. Biological associations can be observed: P1 (Neoplastic) exhibits strong links with G1 (Oncogene) and G2 (Protein Kinase), consistent with the role of oncogenic signaling and kinase activity in tumor proliferation. Similarly, P3 (Inflammatory) shows strong connections with G6 (Cytokines \& Growth Factors), reflecting the involvement of cytokine-mediated signaling in shaping tumor microenvironment and immune responses. 
These findings are consistent with well-established cancer biology \cite{tomuleasa2024therapeutic, lima2021tumor}, suggesting that our BF module not only captures statistical correlations but also reveals the biologically grounded relationships between modalities.

\subsubsection{Multimodal Feature Alignment}
To illustrate the effect of the MA module, Fig.~\ref{fig:vis_tsne} visualizes the evolution of multimodal representations during training. 
At the initial epoch, histology and genomics embeddings are completely separated, indicating weak cross-modal correspondence. 
After applying our MA module, the two modalities become progressively intertwined, revealing improved feature coherence. 
By mid-training (e.g., Epoch 5), embeddings after MA exhibit strong alignment across modalities, suggesting that our MA module effectively bridges the semantic gap between histology and genomics. 
Toward later epochs, both the \textit{before} and \textit{after} alignment representations appear well aligned, implying that the alignment loss has been internalized by the encoders through end-to-end optimization. 
This progression highlights MA as both an early-stage calibrator and a continuing regularizer, promoting consistent multimodal representations across tasks.

\subsubsection{WSI-level Interpretability}
Fig.~\ref{fig:vis_wsi} depicts the spatial attention heatmaps of six histology prototypes on WSIs in survival prediction. Each prototype exhibits distinct spatial distributions. The \textit{Neoplastic} prototype shows strong activation in tumor-dense regions, while the \textit{Stromal} prototypes are more active in stromal areas. The \textit{Inflammatory} prototype highlights immune infiltration, and the \textit{Necrotic} prototype aligns with necrotic cores commonly observed in high-grade gliomas. These spatially coherent activation patterns indicate that the model effectively captures biologically meaningful regions, aligning with known histopathological features. 

\section{Conclusion}
This study presents an interpretable multimodal prototyping framework that integrates histology and genomic data for precision oncology. Our approach jointly addresses four key challenges in multimodal learning: suboptimal WSI representation, ineffective cross-modal alignment, coarse integration, and limited robustness with incomplete genomics. 

Through BP module, the framework embeds domain priors into representation learning, enhancing both clinical relevance and model transparency; the MA strategy enforces semantic consistency through sample-wise and distribution-wise alignment, calibrating the imperfect pairing in multimodal dataset; the BF module captures both shared and complementary signals via cross-modal affinity; and the SGI module reconstructs genomic semantics from histology, ensuring robustness under incomplete or entire missing genomic scenarios. 
Extensive experiments across diagnosis, grading, and survival prediction demonstrate that our framework not only surpasses state-of-the-art models but also remains resilient when genomics are absent. These results highlight that biologically-informed prototyping, semantic alignment, and selective fusion enhance interpretability, robustness, and clinical reliability.

Our work has limitations. 
First, more in-depth analysis of genomics could benefit biological interpretation and relevance. Future efforts will incorporate pathway-level modeling to capture biological interactions. 
Future work will extend to multi-institutional and pan-cancer datasets to enhance clinical translation. Also, although this study focuses on histology and genomics, the proposed framework is readily extensible to additional modalities, such as demographic data, through late-fusion strategies. 
Overall, this study represents a significant step towards developing interpretable, robust, and clinically deployable multimodal learning in precision oncology.

\appendices



\bibliographystyle{IEEEtran}
\bibliography{IEEEabrv,refs}

\end{document}

%% file: tables/table1-2.tex
\begin{table*}[!t]
 \scriptsize
    \setlength{\tabcolsep}{3pt}  
    \caption{Comparison with SOTA methods on Glioma Diagnosis and Grading. 
    p. and g. represent the modality of histology and genomics, respectively.
    Best and second results are highlighted with \textbf{bold} and \underline{underline}.} \label{table1-2}
   \begin{center}
    \resizebox{.99\linewidth}{!}{
      \begin{tabular}{lcc | ccccc | ccccc }
         \toprule
         \multirow{2}{*}{Methods} 
         &\multirow{2}{*}{p.} 
         &\multirow{2}{*}{g.} 
         &\multicolumn{5}{c|}{Diagnosis (\%)}
         &\multicolumn{5}{c}{Grading (\%)}\\ 
         \cmidrule(lr){4-13}           
          &       &       
          & AUC   & Acc.   & Sen.   & Spec.   &F1-score   & AUC   & Acc.   & Sen.   & Spec.   &F1-score  \\
          \midrule
          
         AttMIL~\cite{ilse2018attention}   
         &$\checkmark$     &   
         &$77.26^{\pm 4.28}$ &$53.14^{\pm 3.11}$  &$41.53^{\pm 2.85}$ &$79.32^{\pm 1.24}$ &$32.96^{\pm 4.89}$ 
         &$81.47^{\pm 4.53}$ &$64.29^{\pm 4.33}$  &$51.72^{\pm 2.39}$ &$76.65^{\pm 3.91}$ &$52.86^{\pm 4.75}$\\

         TransMIL~\cite{shao2021transmil}   
         &$\checkmark$     &   
         &$68.28^{\pm 1.98}$ &$43.50^{\pm2.75}$  &$43.60^{\pm 2.73}$ &$77.06^{\pm1.75}$ &$30.14^{\pm 2.03}$
         &$72.88^{\pm 3.24}$ &$57.40^{\pm4.32}$  &$48.09^{\pm 5.83}$ &$75.37^{\pm2.80}$ &$47.94^{\pm 5.14}$\\

         MKD~\cite{zhang2025multi}  
         &$\checkmark$     &   
         &$82.09^{\pm 2.54}$ &$60.37^{\pm3.86}$  &$53.29^{\pm 3.87}$ &$82.01^{\pm1.08}$ &$47.36^{\pm 3.80}$
         &$83.52^{\pm 2.91}$ &$66.78^{\pm2.08}$  &$\mathbf{71.68^{\pm 1.22}}$ &$80.43^{\pm2.07}$ &$64.27^{\pm 3.85}$\\

         Ours (Missing Modality)
         &$\checkmark$     &
         &$\mathbf{86.54^{\pm2.44}}$ &$\mathbf{66.30^{\pm4.36}}$
         &$\mathbf{55.39^{\pm3.45}}$ &$\mathbf{85.41^{\pm1.34}}$ &$\mathbf{53.98^{\pm5.47}}$
         &$\mathbf{87.31^{\pm2.67}}$ &$\mathbf{72.28^{\pm1.67}}$
         &$70.35^{\pm2.38}$ &$\mathbf{82.16^{\pm0.32}}$ &$\mathbf{68.19^{\pm3.77}}$\\

         \midrule
         
         SNN~\cite{klambauer2017self}   
         &   &$\checkmark$   
         &$86.29^{\pm 2.53}$ &$69.13^{\pm6.12}$  &$64.60^{\pm6.25}$ &$88.36^{\pm2.56}$ &$59.53^{\pm 7.26}$
         &$85.70^{\pm 1.88}$ &$73.02^{\pm2.47}$  &$68.19^{\pm4.28}$ &$84.25^{\pm1.95}$ &$66.30^{\pm 3.16}$\\

         Add
         &$\checkmark$     &$\checkmark$
         &$89.68^{\pm 4.17}$ &$68.50^{\pm 2.48}$  &$68.80^{\pm 2.48}$ &$89.26^{\pm 0.27}$ &$57.58^{\pm 2.00}$
         &$87.28^{\pm 2.66}$ &$72.78^{\pm 4.40}$  &$69.42^{\pm 3.88}$ &$80.29^{\pm 1.74}$ &$67.15^{\pm 4.29}$\\

         Concat
         &$\checkmark$     &$\checkmark$
         &$\underline{90.73^{\pm 3.06}}$ &$69.32^{\pm2.13}$  & $65.76^{\pm5.11}$ &$88.95^{\pm3.92}$ &$63.07^{\pm4.57}$
         &$86.69^{\pm 4.52}$ &$69.54^{\pm3.74}$  & $64.20^{\pm3.29}$ &$76.68^{\pm4.02}$ &$61.26^{\pm2.34}$\\
         
         HFBSurv~\cite{li2022hfbsurv}
         &$\checkmark$     &$\checkmark$
         &$85.64^{\pm1.72}$ &$65.20^{\pm3.18}$  &$62.98^{\pm4.53}$ &$86.33^{\pm2.54}$ &$60.42^{\pm6.14}$
         &$83.94^{\pm2.46}$ &$65.98^{\pm4.14}$  &$67.21^{\pm3.28}$ &$79.38^{\pm0.96}$ &$62.37^{\pm4.53}$\\

         MCAT~\cite{chen2021multimodal}
         &$\checkmark$     &$\checkmark$
         &$89.80^{\pm1.81}$ &$72.84^{\pm3.56}$  &$\underline{69.57^{\pm3.06}}$ &$\underline{90.17^{\pm1.29}}$ &$68.86^{\pm2.98}$
         &$\underline{88.26^{\pm1.76}}$ &$\mathbf{75.18^{\pm3.06}}$  &$\underline{72.44^{\pm5.00}}$ &$83.92^{\pm7.13}$ &$65.97^{\pm3.28}$\\ 

         CMTA~\cite{zhou2023cross}
         &$\checkmark$     &$\checkmark$
         &$90.47^{\pm2.39}$ &$\underline{74.50^{\pm2.41}}$  &$67.22^{\pm5.03}$ &$89.64^{\pm1.61}$ &$\underline{71.34^{\pm0.91}}$
         &$87.80^{\pm1.49}$ &$74.74^{\pm1.93}$  &$70.24^{\pm4.89}$ &$\underline{85.02^{\pm1.86}}$ &$\underline{70.25^{\pm3.94}}$\\ 

         MKD~\cite{zhang2025multi}  
         &$\checkmark$     &$\checkmark$   
         &$88.57^{\pm 2.76}$ &$70.34^{\pm3.06}$  &$67.86^{\pm 3.20}$ &$88.15^{\pm2.74}$ &$67.52^{\pm 3.23}$
         &$87.08^{\pm 2.19}$ &$73.56^{\pm2.81}$  &$72.08^{\pm 3.16}$ &$84.29^{\pm2.56}$ &$69.46^{\pm 3.23}$\\

         \midrule
         
         Ours
         &$\checkmark$     &$\checkmark$
         &$\mathbf{92.26^{\pm2.47}}$ &$\mathbf{77.84^{\pm2.31}}$  &$\mathbf{73.89^{\pm1.52}}$ &$\mathbf{92.08^{\pm1.23}}$ &$\mathbf{72.25^{\pm1.74}}$
         &$\mathbf{89.62^{\pm1.37}}$ &$\underline{74.80^{\pm1.18}}$  &$\mathbf{74.29^{\pm1.56}}$ &$\mathbf{87.42^{\pm2.15}}$ &$\mathbf{72.89^{\pm1.57}}$\\ 
         
         \bottomrule
      \end{tabular}
      }
   \end{center}
\end{table*}

%% file: tables/table3.tex
\begin{table}[!t]
 \scriptsize
    \setlength{\tabcolsep}{4pt}  
    \caption{Comparison with SOTA methods on survival prediction. 
    p. and g. represent histology and genomics, respectively. 
    } \label{table3}
   \begin{center}
   \resizebox{.73\linewidth}{!}{
      \begin{tabular}{lcc | cc }
         \toprule
         \multirow{2}{*}{Methods} 
         &\multirow{2}{*}{p.} 
         &\multirow{2}{*}{g.} 
         &\multicolumn{2}{c}{Survival (C-Index, \%)} \\
         \cmidrule(lr){4-5} 
          &       &       
          &  In-domain & Cross-domain \\
          \midrule
          
         AttMIL~\cite{ilse2018attention}   
         &$\checkmark$     &   
         &$70.61^{\pm 3.41}$ &$50.37$\\

         TransMIL~\cite{shao2021transmil}   
         &$\checkmark$     &   
         &$69.56^{\pm 3.02}$&$50.84$\\

         MKD~\cite{zhang2025multi}   
         &$\checkmark$     &   
         &$71.83^{\pm 2.16}$& $49.25$\\

         G-HANet~\cite{wang2025histo}   
         &$\checkmark$     &   
         &$73.72^{\pm 3.24}$&$50.02$\\

         Ours (Missing Modality)   
         &$\checkmark$     &   
         &$\mathbf{75.95^{\pm 2.01}}$&$\mathbf{52.91}$\\

         \midrule         

         SNN~\cite{klambauer2017self}   
         &     &$\checkmark$   
         &$76.46^{\pm 4.32}$&$\underline{62.18}$\\

         Add   
         &$\checkmark$     &$\checkmark$   
         &$78.21^{\pm 2.14}$&$59.20$\\

         Concat   
         &$\checkmark$     &$\checkmark$   
         &$77.10^{\pm 4.26}$&$60.75$\\

         HFBSurv~\cite{li2022hfbsurv}  
         &$\checkmark$     &$\checkmark$   
         &$79.67^{\pm 3.73}$&$61.09$\\

         MCAT~\cite{chen2021multimodal}   
         &$\checkmark$     &$\checkmark$   
         &$\underline{81.80^{\pm 2.56}}$&$53.64$\\
         
         CMTA~\cite{zhou2023cross}   
         &$\checkmark$     &$\checkmark$   
         &$79.24^{\pm 3.75}$&$51.13$\\

         MKD~\cite{zhang2025multi}   
         &$\checkmark$     &$\checkmark$   
         &$80.76^{\pm 3.59}$&$54.67$\\
         
         \midrule

         Ours   
         &$\checkmark$     &$\checkmark$   
         &$\mathbf{83.17^{\pm 2.29}}$&$\mathbf{64.01}$\\

         \bottomrule
      \end{tabular}
      }
   \end{center}
\end{table}

%% file: tables/table4.tex
\begin{table}[!t]
\scriptsize
\setlength{\tabcolsep}{2pt}
\caption{Performance comparison under different types of genomic missingness on survival prediction. 
Best and second results are highlighted with \textbf{bold} and \underline{underline}.}
\label{table4}
\centering
\resizebox{.99\linewidth}{!}{
\begin{tabular}{l|l|ccccc}
\toprule
\multirow{2}{*}{Type} &
\multirow{2}{*}{Methods} &
\multicolumn{5}{c}{C-Index (\%) at different missing rates} \\ 
\cmidrule(lr){3-7}
 & & 100\% & 80\% & 50\% & 20\% & 0\% \\
\midrule

\multirow{4}{*}{Patient-wise}
& Filling   & $58.38^{\pm5.76}$ & $65.38^{\pm3.10}$ & $70.21^{\pm4.61}$ & $74.90^{\pm5.29}$ & \underline{$79.24^{\pm3.75}$} \\
& AE        & $61.16^{\pm5.63}$ & $65.60^{\pm3.17}$ & $72.20^{\pm4.56}$ & \underline{$76.94^{\pm2.76}$} & \underline{$79.24^{\pm3.75}$} \\
& Ensemble  & \underline{$69.43^{\pm4.89}$} & \underline{$70.96^{\pm5.20}$} & \underline{$75.38^{\pm2.14}$} & $76.01^{\pm4.19}$ & $79.02^{\pm3.02}$ \\
& Ours      & $\mathbf{75.95^{\pm2.01}}$ & $\mathbf{76.04^{\pm3.10}}$ & $\mathbf{77.82^{\pm2.35}}$ & $\mathbf{79.38^{\pm3.22}}$ & $\mathbf{83.17^{\pm2.29}}$ \\
\midrule

\multirow{4}{*}{Feature-wise}
& Filling   & $58.38^{\pm5.76}$ & $64.84^{\pm3.63}$ & $74.48^{\pm3.57}$ & $78.90^{\pm2.28}$ & \underline{$79.24^{\pm3.75}$} \\
& AE        & $61.16^{\pm5.63}$ & $69.36^{\pm5.14}$ & $74.03^{\pm3.59}$ & $78.92^{\pm3.10}$ & \underline{$79.24^{\pm3.75}$} \\
& Ensemble  & \underline{$69.43^{\pm4.89}$} & \underline{$70.35^{\pm2.37}$} & \underline{$74.91^{\pm5.28}$} & \underline{$79.15^{\pm3.24}$} & $79.02^{\pm3.02}$ \\
& Ours      & $\mathbf{75.95^{\pm2.01}}$ & $\mathbf{75.94^{\pm4.30}}$ & $\mathbf{77.74^{\pm3.43}}$ & $\mathbf{81.38^{\pm2.95}}$ & $\mathbf{83.17^{\pm2.29}}$ \\
\bottomrule
\end{tabular}
}
\end{table}

%% file: tables/table5.tex
\begin{table*}[!t]
 \scriptsize
 \setlength{\tabcolsep}{3pt}  
    \caption{Ablation Study of Model Components under Missing and Full Genomic Conditions. Best results are highlighted with \textbf{bold}.} \label{table5}
   \begin{center}
   \resizebox{.78\linewidth}{!}{
      \begin{tabular}{l|l|ccc|ccc}
         \toprule
          \multirow{2}{*}{Module} & \multirow{2}{*}{Variant} & \multicolumn{3}{c|}{Missing Modality (\%)} & \multicolumn{3}{c}{Full Modality (\%)}\\
        \cmidrule(lr){3-8}
        & & Diagnosis (Acc.)&Grading (Acc.) & Survival (C-Index)&Diagnosis (Acc.)&Grading (Acc.)&Survival (C-Index)\\
         \midrule
        \multirow{3}{*}{BP} & w/o BP (Random Init) & $63.57^{\pm 4.51}$ & $69.05^{\pm 2.93}$ & $74.19^{\pm 5.96}$ & $75.01^{\pm 0.19}$& $70.58^{\pm 2.81}$& $80.17^{\pm 3.77}$\\
          & Ours &  $\mathbf{67.52^{\pm 3.94}}$ & $\mathbf{72.30^{\pm 2.84}}$ & $\mathbf{75.95^{\pm 2.01}}$ & $\mathbf{77.84^{\pm 2.31}}$ & $\mathbf{74.83^{\pm 2.67}}$ & $\mathbf{83.17^{\pm 2.29}}$\\
         \midrule
         
        \multirow{4}{*}{MA} & w/o MA & $45.69^{\pm 5.48}$ & $58.42^{\pm 2.92}$ & $65.48^{\pm 3.19}$ & $65.48^{\pm 3.19}$ & $66.27^{\pm 2.64}$ & $78.33^{\pm 4.56}$\\
        & w/o Sample-wise Alignment & $62.42 ^{\pm 4.08 }$ & $69.88^{\pm 2.43}$ &$72.15^{\pm 2.52}$& $73.58^{\pm 3.98}$ & $72.96^{\pm 1.84}$ &$80.59^{\pm 3.92}$\\
         & w/o Distribution-wise Alignment & $63.59^{\pm 3.97}$ & $69.72^{\pm 3.07}$ &$69.55^{\pm 4.18}$& $74.62^{\pm 2.73}$ & $71.50^{\pm 2.87}$ & $81.36^{\pm 5.02}$\\  
          & Ours &  $\mathbf{67.52^{\pm 3.94}}$ & $\mathbf{72.30^{\pm 2.84}}$ & $\mathbf{75.95^{\pm 2.01}}$ & $\mathbf{77.84^{\pm 2.31}}$ & $\mathbf{74.83^{\pm 2.67}}$ & $\mathbf{83.17^{\pm 2.29}}$\\
          \midrule
          
          \multirow{4}{*}{SGI} & w/o SGI & \textemdash & \textemdash &\textemdash & $76.28^{\pm 1.81}$ & $73.25^{\pm 2.19}$ &$82.02^{\pm 2.39}$\\
          & Diffusion-based Imputation & $48.71^{\pm 5.28}$ & $50.47^{\pm 3.84}$ &$60.33^{\pm 6.29}$& $76.28^{\pm 1.81}$ & $73.25^{\pm 2.19}$ &$82.02^{\pm 2.39}$\\
          & GAN-based (w/o Interpolation) & $60.85^{\pm 3.83}$ & $69.46^{\pm 2.53}$ &$72.06^{\pm 3.30}$ & $76.80^{\pm 3.82}$ & $73.02^{\pm 1.47}$ &$82.93^{\pm 3.10}$\\
          & Ours &  $\mathbf{67.52^{\pm 3.94}}$ & $\mathbf{72.30^{\pm 2.84}}$ & $\mathbf{75.95^{\pm 2.01}}$ & $\mathbf{77.84^{\pm 2.31}}$ & $\mathbf{74.83^{\pm 2.67}}$ & $\mathbf{83.17^{\pm 2.29}}$\\
          \midrule
          
           \multirow{2}{*}{BF} & w/o BF & $63.52^{\pm 5.46}$ & $70.33^{\pm 2.56}$ &$73.42^{\pm 3.67}$ & $74.09^{\pm 2.83}$ & $72.56^{\pm 2.43}$ & $82.56^{\pm 2.00}$\\
           & Ours &  $\mathbf{67.52^{\pm 3.94}}$ & $\mathbf{72.30^{\pm 2.84}}$ & $\mathbf{75.95^{\pm 2.01}}$ & $\mathbf{77.84^{\pm 2.31}}$ & $\mathbf{74.83^{\pm 2.67}}$ & $\mathbf{83.17^{\pm 2.29}}$\\

         \bottomrule
      \end{tabular}
      }
   \end{center}
\end{table*}

%% file: tables/table6.tex
\begin{table}[t]
\setlength{\tabcolsep}{3pt}  
\caption{Hyper-parameter analysis on the Top-$K$ parameter in the BF module for survival prediction. (C-index, \%)}
\label{tab:ablate_k_survival}
\centering
\small
\resizebox{.95\linewidth}{!}{
\begin{tabular}{l|cccccccc}
\toprule
Metric & $K=0$ & $K=1$ & $K=2$ & $K=3$ & $K=4$ & $K=5$ & $K=6$ \\
\midrule
In-domain $\uparrow$     & 82.56 & 82.43 & 83.09 & \textbf{83.17} & 82.91 & 82.62 & 82.35 \\
Cross-domain $\uparrow$  & 63.19 & 63.22 & \textbf{64.88} & 64.01 & 63.73 & 63.44 & 63.17 \\
Train time (h) $\downarrow$ & 3.83  & 3.77  & 3.71  & 3.67  & 3.62  & 3.56  & 3.53 \\
\bottomrule
\end{tabular}
}
\end{table}